%% file: iclr2021_conference.tex
\documentclass{article} 
\usepackage{iclr2021_conference,times}

\input{math_commands.tex}

\usepackage{hyperref}
\usepackage{url}

\usepackage{graphicx}
\usepackage{algorithm}
\usepackage{algorithmic}
\usepackage{caption}
\usepackage{subcaption}

\title{Understanding the Effect of the Long Tail on Neural Network Compression}

\author{Harvey Dam, Aditya Bhaskara \& Ganesh Gopalakrishnan \\
School of Computing, University of Utah\\
Salt Lake City, UT, USA \\
\texttt{harvey,aditya,ganesh@cs.utah.edu}\\
\AND
Vinu Joseph, Saurav Muralidharan \& Michael Garland \\
NVIDIA \\
Santa Clara, CA, USA \\
\texttt{vinuj,sauravm,michael@nvidia.com}
}

\iclrfinalcopy 
\begin{document}

\maketitle
\input{ICLR-SNN/tex/abstract}
\input{ICLR-SNN/tex/intro}

\input{ICLR-SNN/tex/methodology}
\input{ICLR-SNN/tex/results}
\newpage

\bibliography{iclr2021_conference}
\bibliographystyle{iclr2021_conference}

\section{Appendix}
\input{ICLR-SNN/tex/appendix.tex}

\end{document}

%% file: math_commands.tex
\usepackage{amsmath,amsfonts,bm}

\def\eqref#1{equation~\ref{#1}}

\def\1{\bm{1}}

\DeclareMathAlphabet{\mathsfit}{\encodingdefault}{\sfdefault}{m}{sl}
\SetMathAlphabet{\mathsfit}{bold}{\encodingdefault}{\sfdefault}{bx}{n}



%% file: ICLR-SNN/tex/abstract.tex
\begin{abstract}
Network compression is now a mature sub-field of neural network research: over the last decade, significant progress has been made towards reducing the size of models and speeding up inference, while maintaining the classification accuracy. However, many works have observed that focusing on just the overall accuracy can be misguided. E.g., it has been shown that mismatches between the full and compressed models can be biased towards under-represented classes. This raises the important research question, \emph{can we achieve network compression while maintaining ``semantic equivalence'' with the original network?} In this work, we study this question in the context of the ``long tail'' phenomenon in computer vision datasets observed by~\cite{feldman2020does}. They argue that \emph{memorization} of certain inputs (appropriately defined) is essential to achieving good generalization. As compression limits the capacity of a network (and hence also its ability to memorize), we study the question: are mismatches between the full and compressed models correlated with the memorized training data? We present positive evidence in this direction for image classification tasks, by considering different base architectures and compression schemes.   
%
%
\end{abstract}

%% file: ICLR-SNN/tex/intro.tex
\section{Introduction}

A large body of research has been devoted to developing methods that can reduce the size of deep neural network (DNN) models considerably without affecting the standard metrics such as top-1 accuracy.
Despite these advances, there are still {\em mismatches} between the models, i.e., inputs that are classified differently by the original and compressed models. Furthermore, there has been evidence that compression can affect the accuracy of certain classes more than others (leading to fairness concerns, e.g., see \cite{hooker2019compressed, joseph2020correctness}).
%

Many works have tried to combat mismatches by using techniques such as reweighting misclassified examples~\cite{focalloss}, using multi-part loss functions~\cite{joseph2020going} that incorporate ideas from knowledge-distillation~\cite{hinton2015distilling} to induce better \emph{alignment} between the original and compressed models. 
\cite{DBLP:phd/us/Joseph21} demonstrated that inducing alignment also improves metrics such as fairness across classes and similarity of attribution maps. 
In spite of this progress, all the known techniques for model compression result in a non-negligible number of mismatches. This leads to some natural questions, {\em can we develop a systematic understanding about these mismatches? Are a certain number of mismatches unavoidable and an inherent consequence of underparameterization?} 
These questions are important not only for vision models, but also in other domains such as language models~\cite{Brown2020LanguageAI}, autonomous driving~\cite{bojarski2016end},
health-care~\cite{ravi2016deep} and
finance~\cite{tran2020deep}.

\begin{figure}
     \centering
     \begin{subfigure}[b]{0.32\textwidth}
         \centering
         \includegraphics[width=\textwidth]{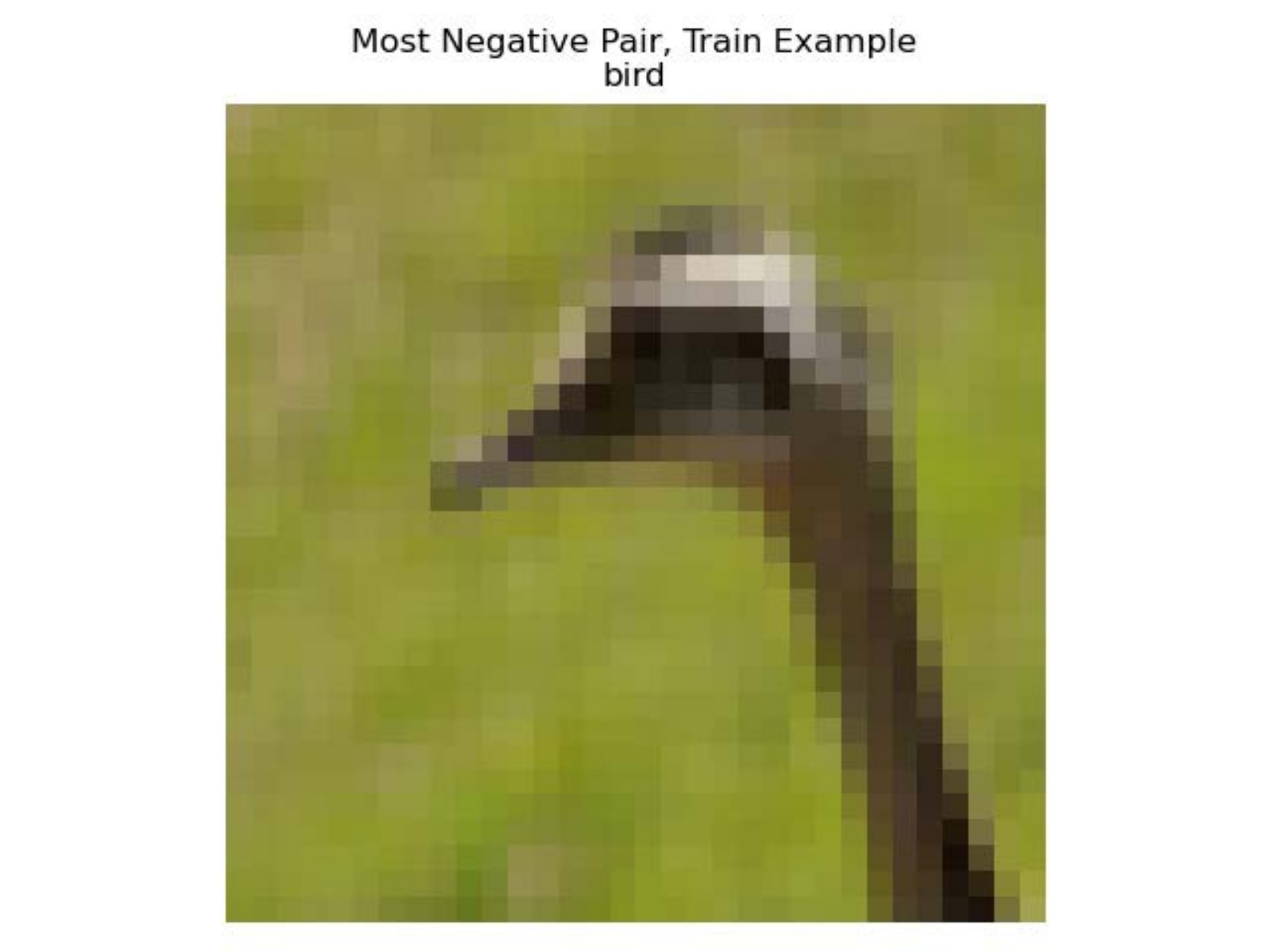}
         \caption{}
     \end{subfigure}
     \hfill
     \begin{subfigure}[b]{0.32\textwidth}
         \centering
         \includegraphics[width=\textwidth]{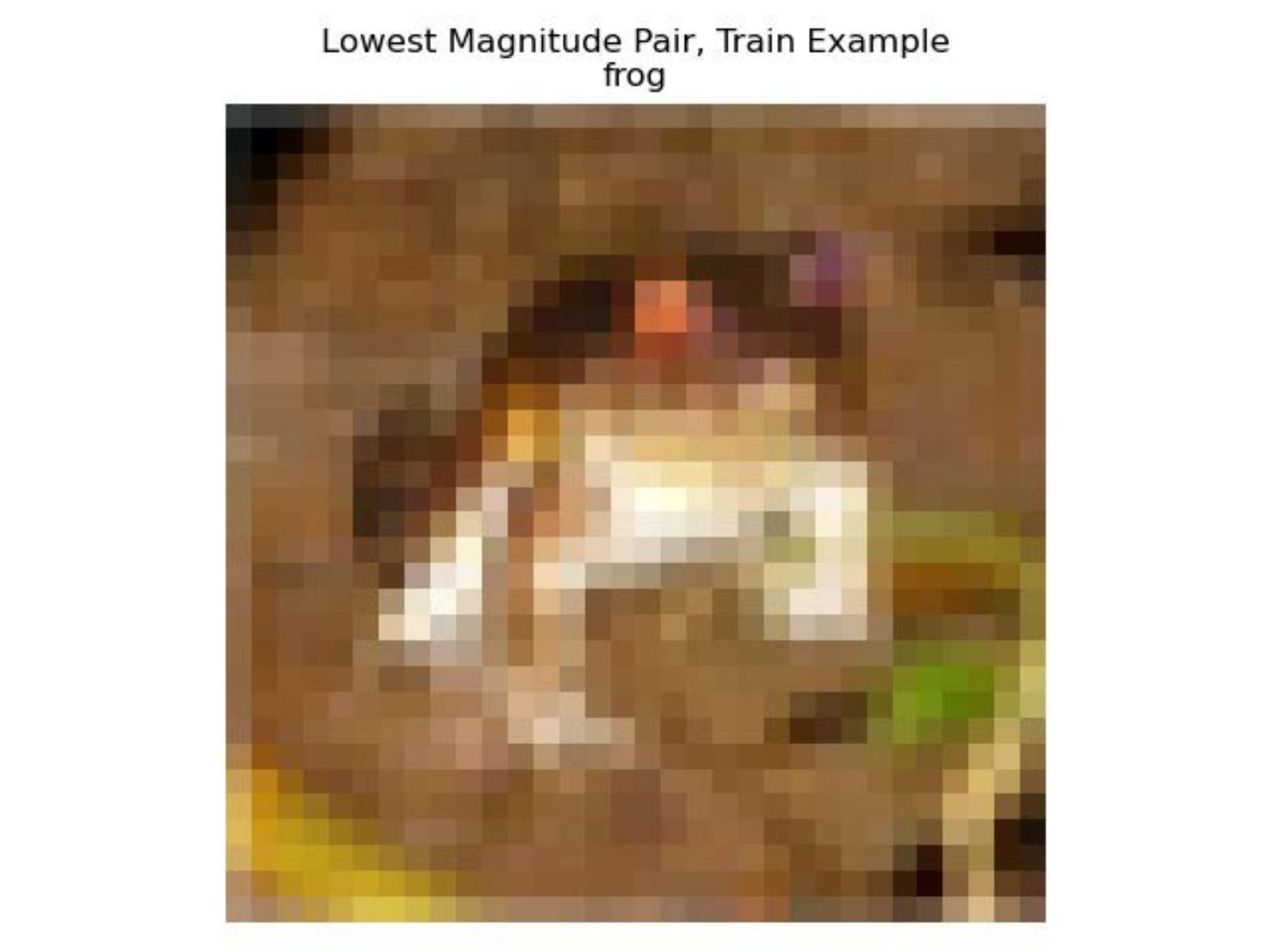}
         \caption{}
     \end{subfigure}
    \hfill
    \begin{subfigure}[b]{0.32\textwidth}
         \centering
         \includegraphics[width=\textwidth]{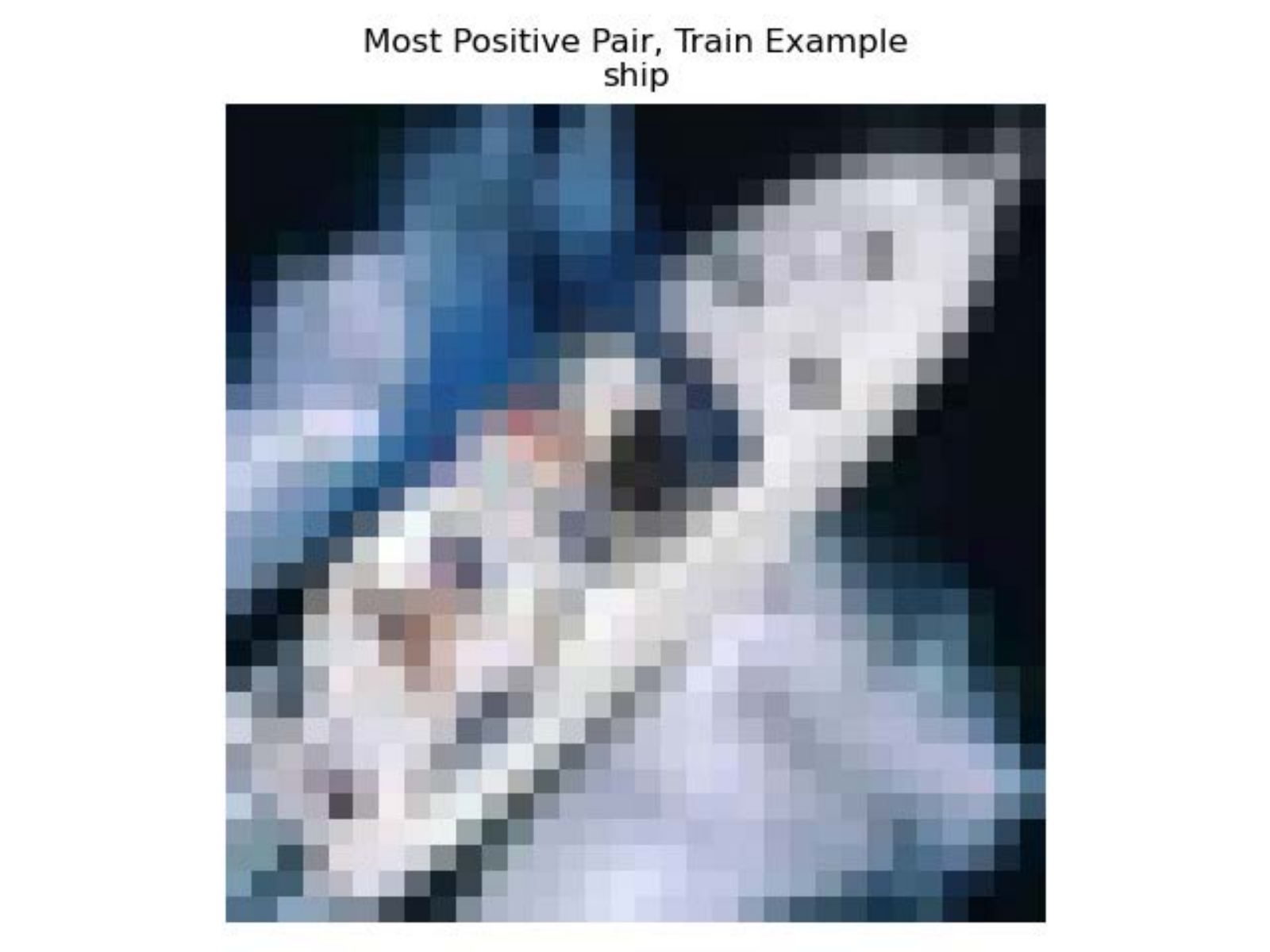}
         \caption{}
    \end{subfigure}
     \hfill
     \begin{subfigure}[b]{0.32\textwidth}
         \centering
         \includegraphics[width=\textwidth]{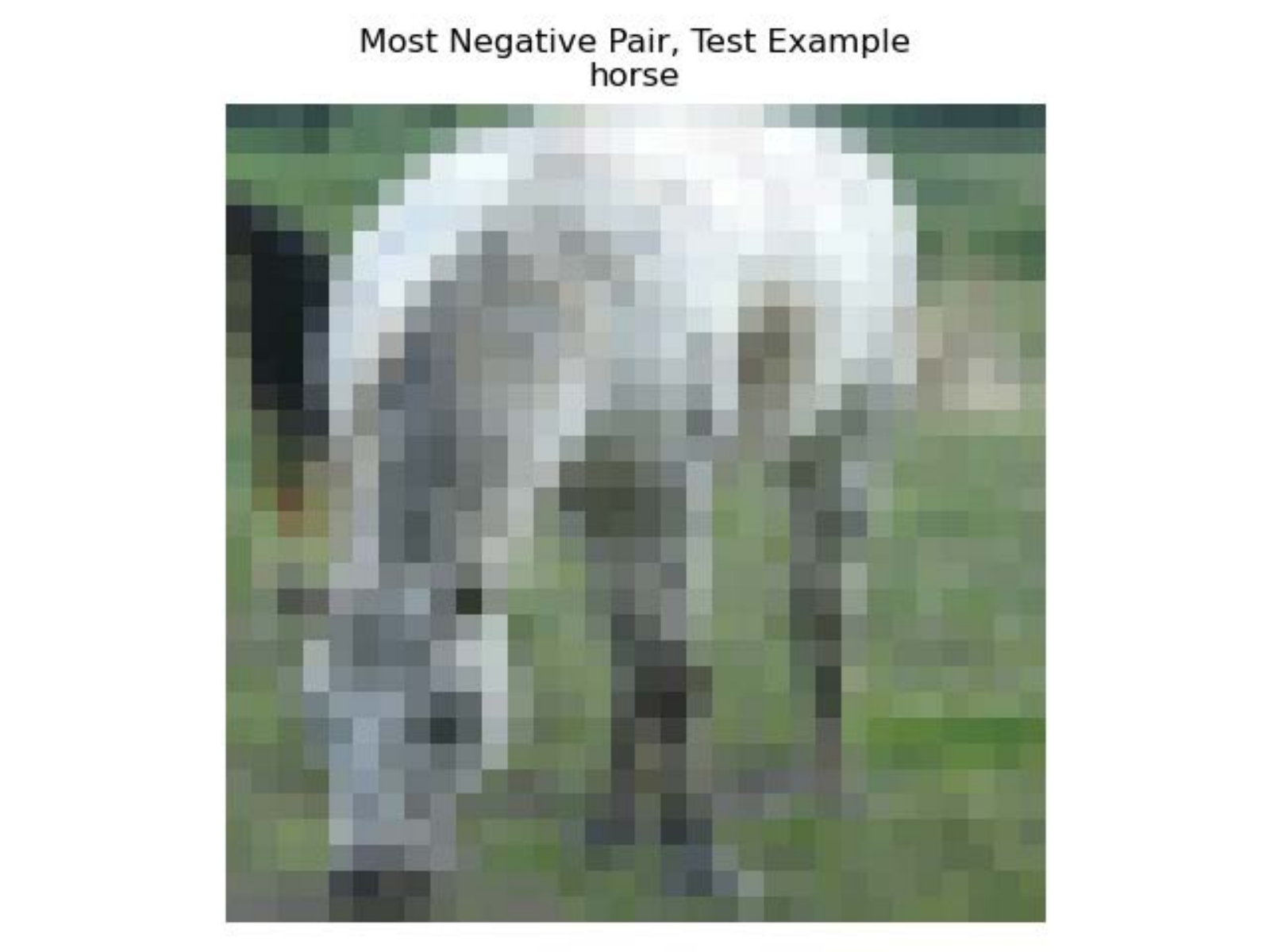}
         \caption{}
     \end{subfigure}
     \hfill
     \begin{subfigure}[b]{0.32\textwidth}
        \centering
        \includegraphics[width=\textwidth]{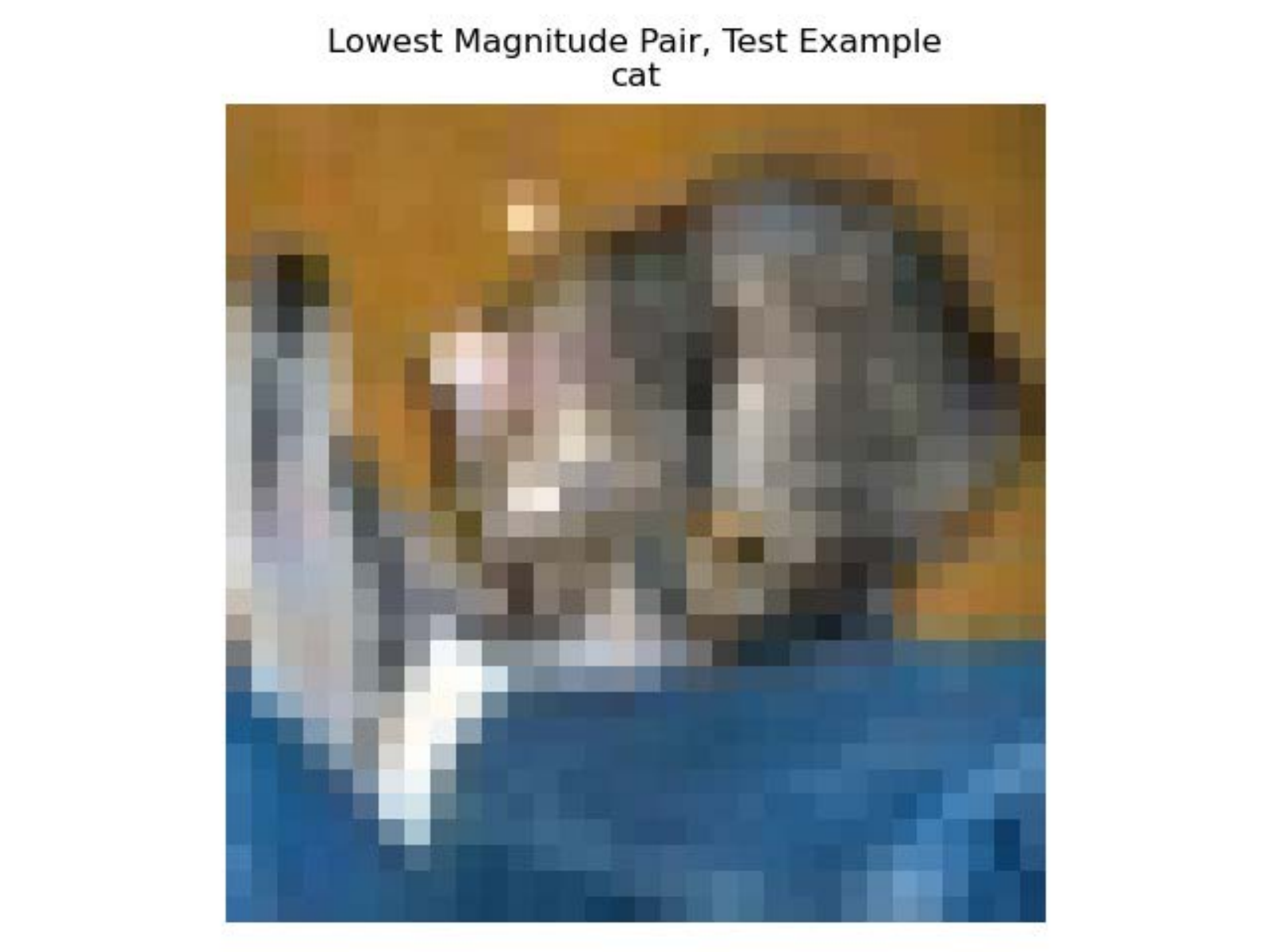}
        \caption{}
    \end{subfigure}
    \hfill
    \begin{subfigure}[b]{0.32\textwidth}
         \centering
         \includegraphics[width=\textwidth]{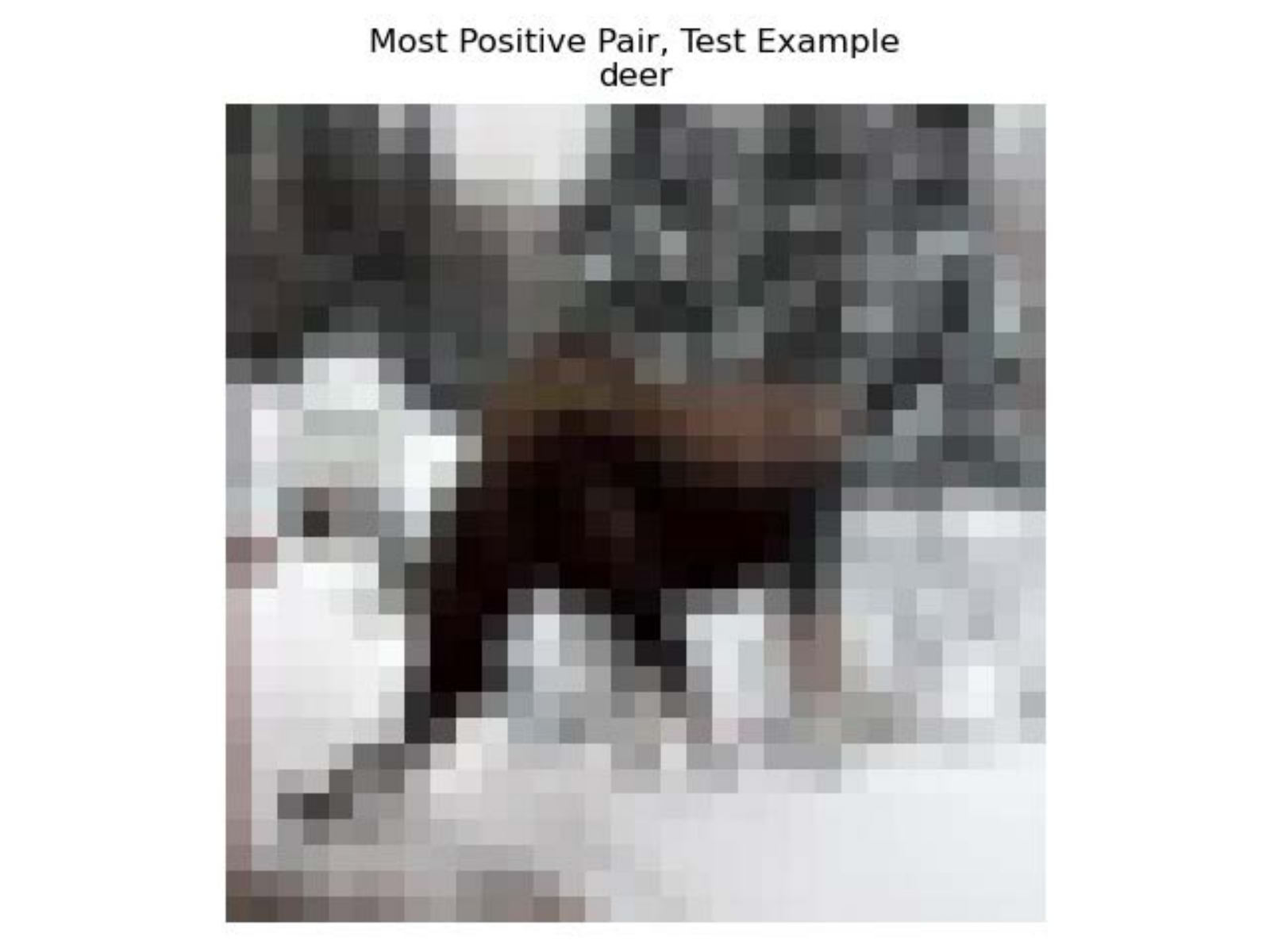}
         \caption{}
    \end{subfigure}
    \caption{CIFAR10 examples corresponding to extreme influence values. (a, d) Train and test examples at the most negative influence value, (b, e) at the lowest magnitude influence value, and (c, f) at the most positive influence value. In general, we expect positive influence to correspond to \emph{helpful} examples, near-zero influence to correspond to unremarkable examples, and negative influence to correspond to unhelpful examples.}
    \label{fig:exampleimages}
\end{figure}

Our goal in this work is to address these questions using the idea that real datasets have a ``long tail'' property, as hypothesized in~\cite{feldman2020does, feldman2020neural}. Informally, their thesis is that real datasets have many examples that are ``atypical,'' and unless their labels are \emph{memorized}, we incur high test loss. Figure~\ref{fig:exampleimages} shows some such atypical examples. This thesis has a direct implication for network compression: if the compressed network does not have sufficient capacity to memorize the atypical examples, it must have a significant mismatch with the original model! Our goal in this work is to explore this connection by asking,    
%
%
%
{\em is there a strong correlation between the mismatches that arise after model compression, and the long-tail portion of the data distribution?}

To answer this question, we first compute the \emph{influence}, as defined by~\cite{feldman2020neural}, of each training example on the model's accuracy on test examples, as well as on training examples, in the CIFAR-10 dataset \cite{Krizhevsky09learningmultiple}. We then compress models using a variety of compression algorithms and perform statistical analysis on the influence values and mismatched predictions between the reference model and the compressed model. Our experiments are based on the observation that if training and test sets are drawn from the same distribution, then memorizing highly influential, atypical, training examples improves accuracy on test examples that are similar to them and also atypical. By observing the connection between these highly influenced test examples and a compressed model's misclassified test examples, we can characterize the misclassified examples in terms of memorization.

%% file: ICLR-SNN/tex/methodology.tex
\section{Definitions and Methodology}

\subsection{Measuring Mismatch or Misalignment}
\label{sec:metrics_misalignment}

\begin{figure}
    \centering
    \includegraphics[width=\textwidth]{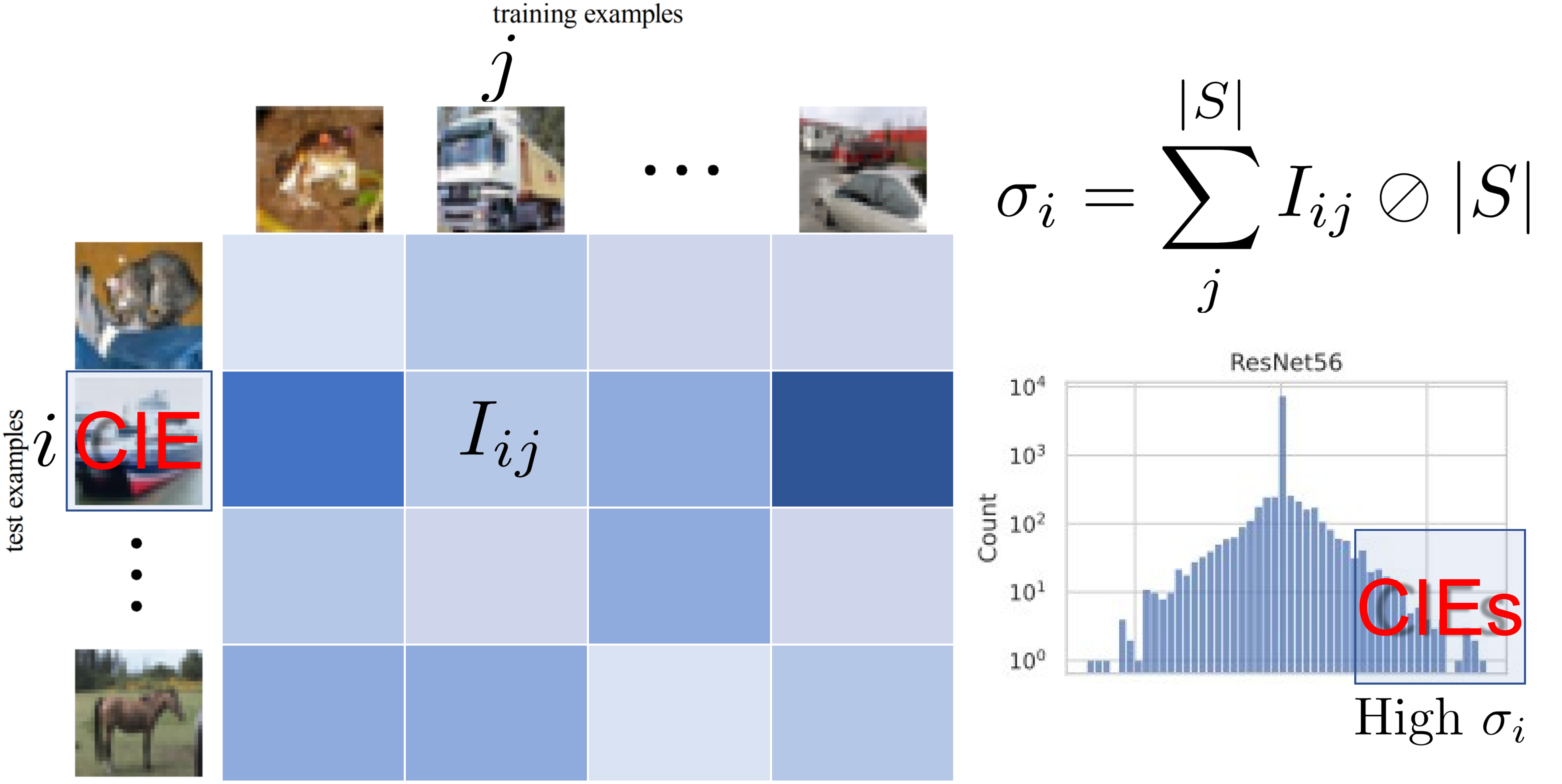}
    \caption{We collect the estimated influence of each training example on each test example is collected in a matrix $I$, shaped (number of test examples $\times$ number of training examples). The $(i, j)$th element is the estimated influence of training example $j$ on test example $i$.}
    \label{fig:influencmatrixaverages}
\end{figure}

Comparing DNN models is a well-known challenge in machine learning (ML) systems. In the context of model compression, it is natural to compare models in terms of their ``functional'' behavior. This leads to the definition of \textit{compression impacted exemplars} presented below. We remark that other metrics are also very useful, e.g., understanding whether the models ``use the same features'' for classification; obtaining concrete metrics that can capture such semantic information is an active direction of research. 

The simplest measure of misalignment among models is in terms of their predictions. \cite{hooker2019compressed} introduced the notion of \textbf{compression impacted exemplars (CIEs)}: test examples that the compressed and uncompressed models classify {\em differently}. Having a near-zero number of CIEs indicates that the models are ``functionally'' equivalent. We will focus on the CIEs in our experiments, and also divide them into subtypes. Note that CIEs are more important when the reference model makes the {\em correct} (agreeing with ground truth) prediction, because this means that the compressed models get those examples wrong. We denote such examples by {\em CIE-U}. 
CIEs that are classified correctly by the compressed model are denoted {\em CIE-C}.


\subsection{Influence and Memorization of Training Data}

As outlined earlier, \cite{feldman2020does} makes the case that memorization is \emph{necessary} for achieving close-to-optimal generalization error in real datasets. Specifically, when the data distribution is long-tailed, i.e., when rare and atypical instances make up a significant fraction of the data, memorizing these instances is unavoidable in order to obtain high accuracy. 
%
%

\cite{feldman2020neural} developed approaches for empirically evaluating this ``long tail'' theory.
The starting point for such evaluation is examining which training examples are memorized and the utility of the memorized examples as a whole.
We recall their definitions of \textit{influence} and \textit{memorization}:
For a training algorithm $\mathcal{A}$ 
operating on a training dataset $S = ((x_1,y_1),\cdots,(x_{|S|},y_{|S|}))$
and test dataset $T = ((x'_1,y'_1),\cdots,(x'_{|T|},y'_{|T|}))$
the amount of influence of $(x_i,y_i) \in S$ on $(x'_j, y'_j) \in T$
is the difference between the accuracy of classifying $(x'_j, y'_j)$ 
after training with and without $(x_i,y_i)$. Formally,

\begin{equation}
\texttt{infl}(\mathcal{A}, S, i, j):= 
 \underset{h \leftarrow \mathcal{A}(S)}{\mathbf{Pr}}\left[h\left(x'_j\right)=y'_j\right] \\ 
-   \underset{h \leftarrow \mathcal{A}(S^{\backslash i})}{\mathbf{Pr}}\left[h\left(x'_j\right)=y'_j\right]
\end{equation}

where $S^{\backslash i}$ denotes the dataset $S$ with $(x_i,y_i)$
removed and probability is taken over the randomness of the 
algorithm $\mathcal{A}$ such as random initialization.
Memorization is defined as the influence of training examples on training examples, i.e. when $S = T$ and $(x_i, y_i) = (x'_j, y'_j)$. 
This definition captures and quantifies the intuition that 
an algorithm memorizes the label $y_i$ if its prediction
at $x_i$ based on the rest of the dataset changes significantly
once $(x_i,y_i)$ is added to the dataset.

Using the definition directly, calculating the influence requires training $\mathcal{A}(S^{\backslash i})$ on the order of $1/\sigma^2$ times for every example, $\sigma^2$ being the variance of the estimation.
As a result, this approach requires $\Omega(|S|/\sigma^2)$ training runs which translates into millions of training runs needed to achieve $\sigma \le 0.1$
on a dataset with $|S| = 50,000$ examples.
To avoid this blow-up,~\cite{feldman2020neural} provide an estimation algorithm that uses only $O(1/\sigma^2)$ training steps in total. 
We adopt their algorithm in our experiments.

\section{Experiments}

\begin{figure}
    \centering
    \includegraphics[width=\textwidth]{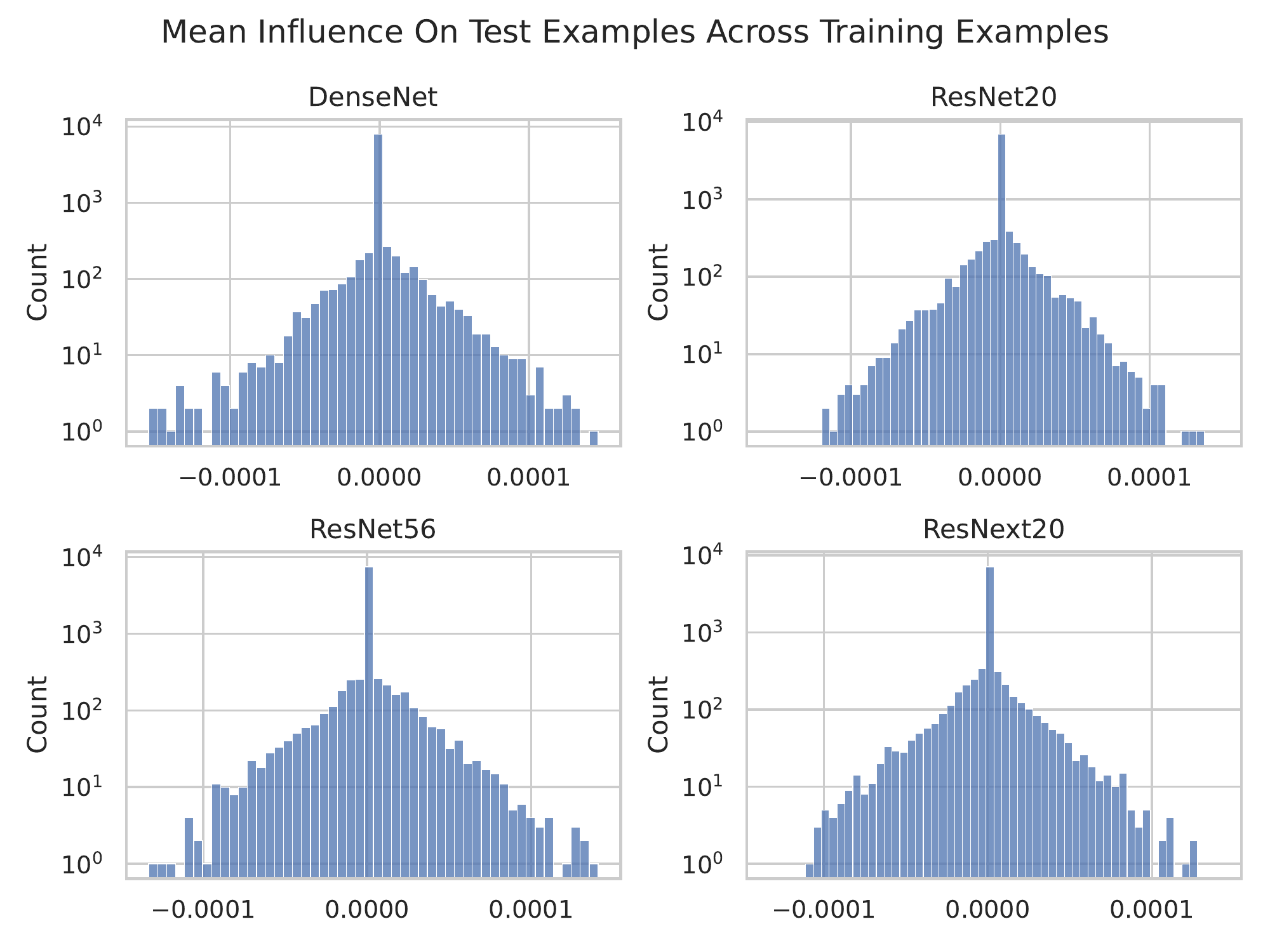}
    \caption{Histograms of estimated mean influence on CIFAR10 test examples across training examples, i.e. the distribution of $(\sum_j I_{:,j}) \oslash |S|$. Note the logarithmic vertical scale. Almost all training examples have close to 0 influence across the test set. The minority that influences the test set at all do not affect accuracy more than about $10^{-4}$.}
    \label{fig:cifar10influenced}
\end{figure}

Although individual influence values are bounded in $[-1, 1]$, influence values on examples from natural data tend to be roughly normally distributed and close to zero (Figure~\ref{fig:cifar10influenced}). Furthermore, there are usually much fewer CIEs than non-CIEs in a test set. Thus, using the Student's t-test to compare the influences of different sets of examples is reasonable. Using the estimated influence and memorization of CIFAR-10 training examples from \cite{feldman2020does}, along with the CIEs that resulted from compression, we compared the mean influence on CIEs versus those on examples that were not CIEs. A t-test was done for each loss functions and CIE types. These experiments are summarized in Algorithm~\ref{alg:influenceestimationttest}, in which $\odot$ is element-wise multiplication and $\oslash$ is element-wise division. 

\begin{figure}[h]
    \centering
    \includegraphics[width=\textwidth]{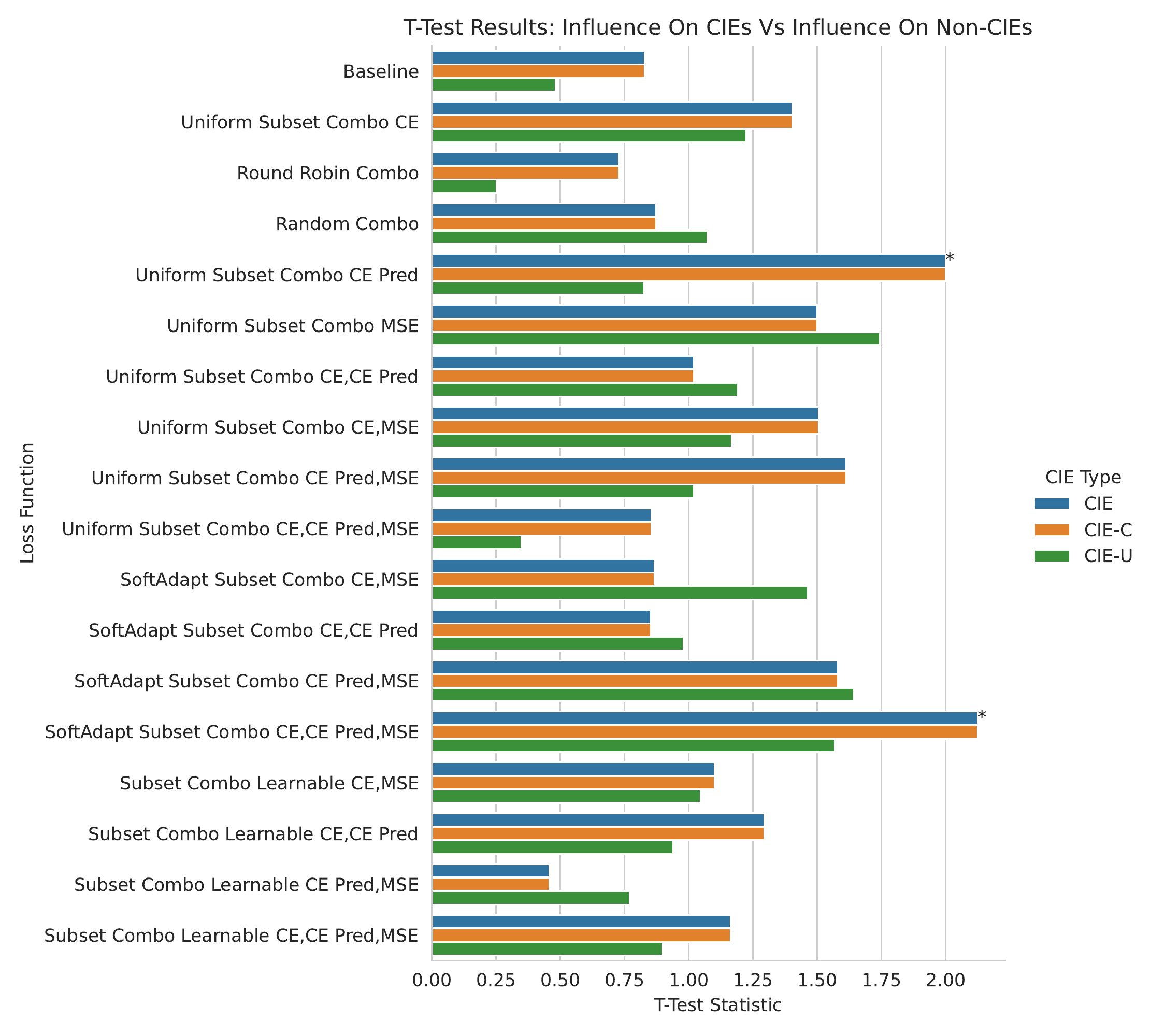}
    \caption{
    T-test results between the mean difference between influence on CIEs and on non-CIEs. Tests that were significant with p-value $\leq 0.05$ are marked with *. The t-statistic reflects the number of CIEs that resulted from compression, with the loss functions that resulted in fewer CIEs exhibiting less influence magnitude among them.}
    \label{fig:cieinfluencedifference}
\end{figure}

We implemented Algorithm~\ref{alg:influenceestimationttest} in Python using PyTorch 1.13 and estimated the influence and memorization of CIFAR10 examples in DenseNet, ResNet-20, ResNet-56, and ResNext-20 over 100 trials each. For each trial, we masked out a random 30\% of the training set and trained for 300 epochs, after which we evaluated correctness on the model using parameters that achieved the highest test accuracy. We compressed ResNet-56 using parameterized loss functions introduced in \cite{joseph2020going} and identified the CIEs that resulted from each loss function.

%% file: ICLR-SNN/tex/results.tex

Our main results are in Figure~\ref{fig:cieinfluencedifference}, which shows the test statistics that resulted from performing the t-test using the baseline loss function in Group Sparsity and using the parameterized loss functions in \cite{joseph2020going}. (We report CIE counts in Figure~\ref{fig:ciescount}.) Compressing while using any loss function tested resulted in some positive test statistic, with two loss functions resulting in p-value $\leq 0.05$. One of those, SoftAdapt \cite{softadapt}, also happened to achieve the lowest number of CIEs. In general, this indicates that CIEs tend to be unusually highly influenced, and expected to be similar to highly influential, atypical training examples.

\section{Conclusions}

This paper has presented a novel method for understanding the residual impact of label preserving neural network compression. We have discovered that the label mismatches of the compressed model occurred where the classification accuracy of data points were most affected by the presence or absence of training examples.  
With this initial understanding in place, we envision a number of directions to expand on this idea by confirming this behavior on larger datasets such as ImageNet \cite{imagenet} and extending this to large language models' \cite{OpenAI} memorization of irrelevant text data such as social security numbers. 
We intend to extend this analysis to help debug compression (sparsity and quantization) induced classification mismatches in privacy preserving inference frameworks that we have recently developed \cite{gouert2023arctyrex, gouert2023accelerated}.

%% file: ICLR-SNN/tex/appendix.tex
\subsection{Algorithms}


\begin{algorithm}[H]
    \caption{Model Compression Mismatch Analysis using Influence and Memorization Estimation}
    \label{alg:influenceestimationttest}
    \begin{algorithmic}
        \STATE \textbf{Require:} classifier $\mathcal{A}$, training set $S = (X, Y)$, test set $T = (X', Y')$, number of samples $t$, proportion of training set to sample $p$
        \STATE \textbf{Return:} estimated influence $I$ where $I_{ij}$ is the influence of training example $j$ on test example $i$, estimated memorization $M$ where $M_{ij}$ is the influence of training example $j$ on training example $i$, test statistic and p-value of a t-test between CIEs and non-CIEs
        \STATE Initialize array $\mathbf{M}_{t \times |X|}$ to $\mathbf{0}$ \hfill \COMMENT{masks}
        \STATE Initialize array $\mathbf{C}_{t \times |Y|}$ \hfill \COMMENT{train correctness}
        \STATE Initialize array $\mathbf{D}_{t \times |Y'|}$ \hfill \COMMENT{test correctness}
        \FOR{$k \gets 1$ to $t$}
            \STATE $\mathbf{M}_{kj} \gets$ 1 with probability $p$ for all $j$
            \STATE Train $\mathcal{A}$ on $S_{\mathbf{M}_k}$
            \STATE $\mathbf{C}_k \gets \mathcal{A} (X) = Y$
            \STATE $\mathbf{D}_k \gets \mathcal{A} (X') = Y'$
        \ENDFOR
        \STATE $I \gets \mathbf{D}^\mathsf{T} \mathbf{M} \oslash \sum_k \mathbf{M}_k - \mathbf{D}^\mathsf{T} \neg \mathbf{M} \oslash \sum_k \neg \mathbf{M}_k$  
        \STATE $M \gets \mathbf{C} \odot \mathbf{M} \oslash \sum_k \mathbf{M}_k - \mathbf{C} \odot \neg \mathbf{M} \oslash \sum_k \neg \mathbf{M}_k$
        \STATE $\bar{\mathcal{A}} \gets$ compress $\mathcal{A}$
        \STATE $\texttt{CIE} \gets \mathcal{A} (X') \neq \bar{\mathcal{A}} (X')$ \hfill \COMMENT{mismatches}
        \STATE test statistic, p-value $\gets$ t-test between $((\sum_j I_{:,j}) \oslash |S|)_\texttt{CIE}$ and $((\sum_j I_{:,j}) \oslash |S|)_{\neg \texttt{CIE}}$
    \end{algorithmic}
\end{algorithm}

\subsection{Additional figures}

\begin{figure}[h]
    \centering
    \includegraphics[width=0.8\textwidth]{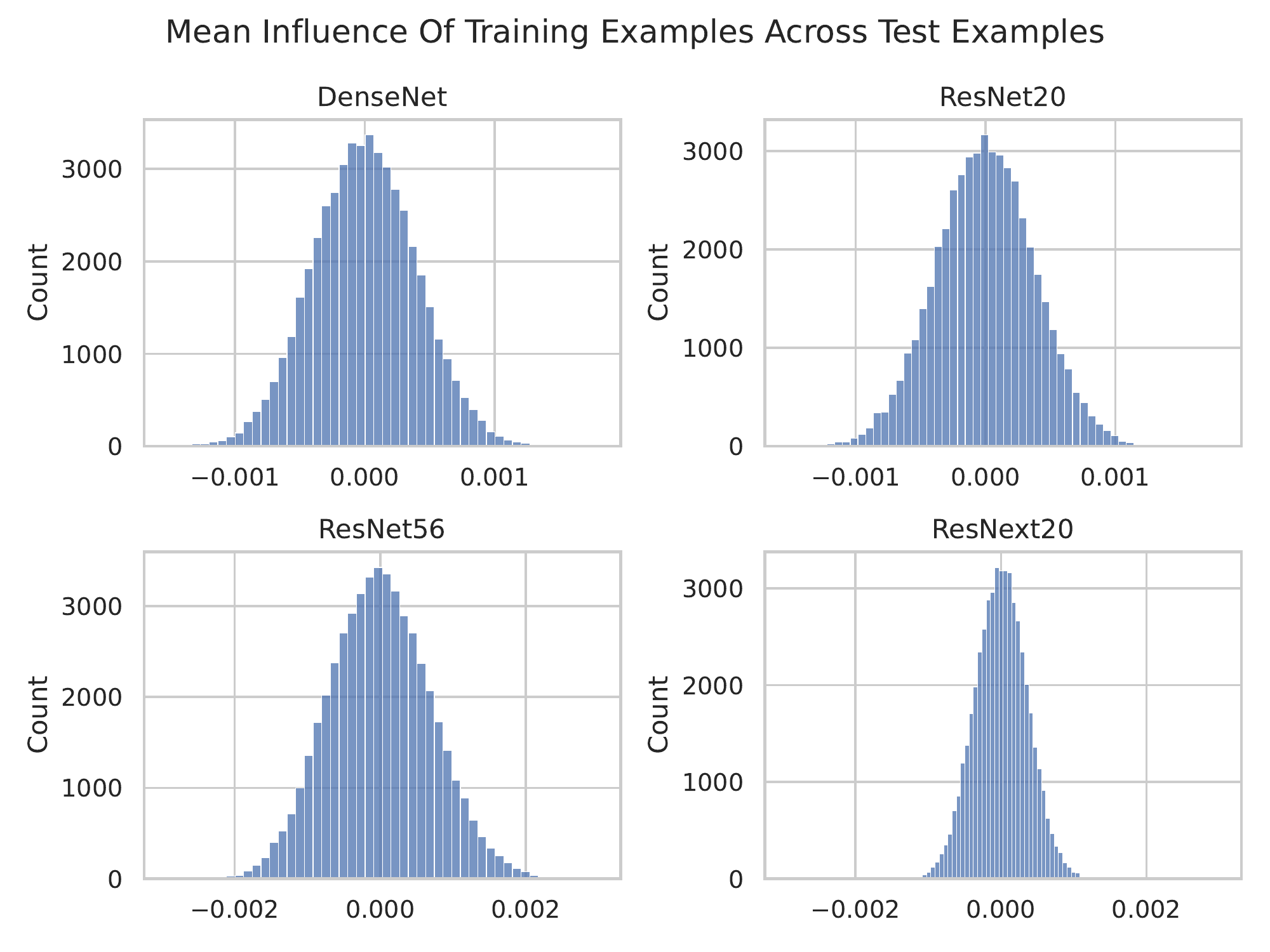}
    \caption{Histograms of estimated mean influence of CIFAR10 training examples across test examples, i.e. the distribution of $(\sum_i I_{i,:}) \oslash |T|$. The horizontal axes show influence values.}
    \label{fig:cifar10influence}
\end{figure}

\begin{figure}[h]
    \centering
    \includegraphics[width=\textwidth]{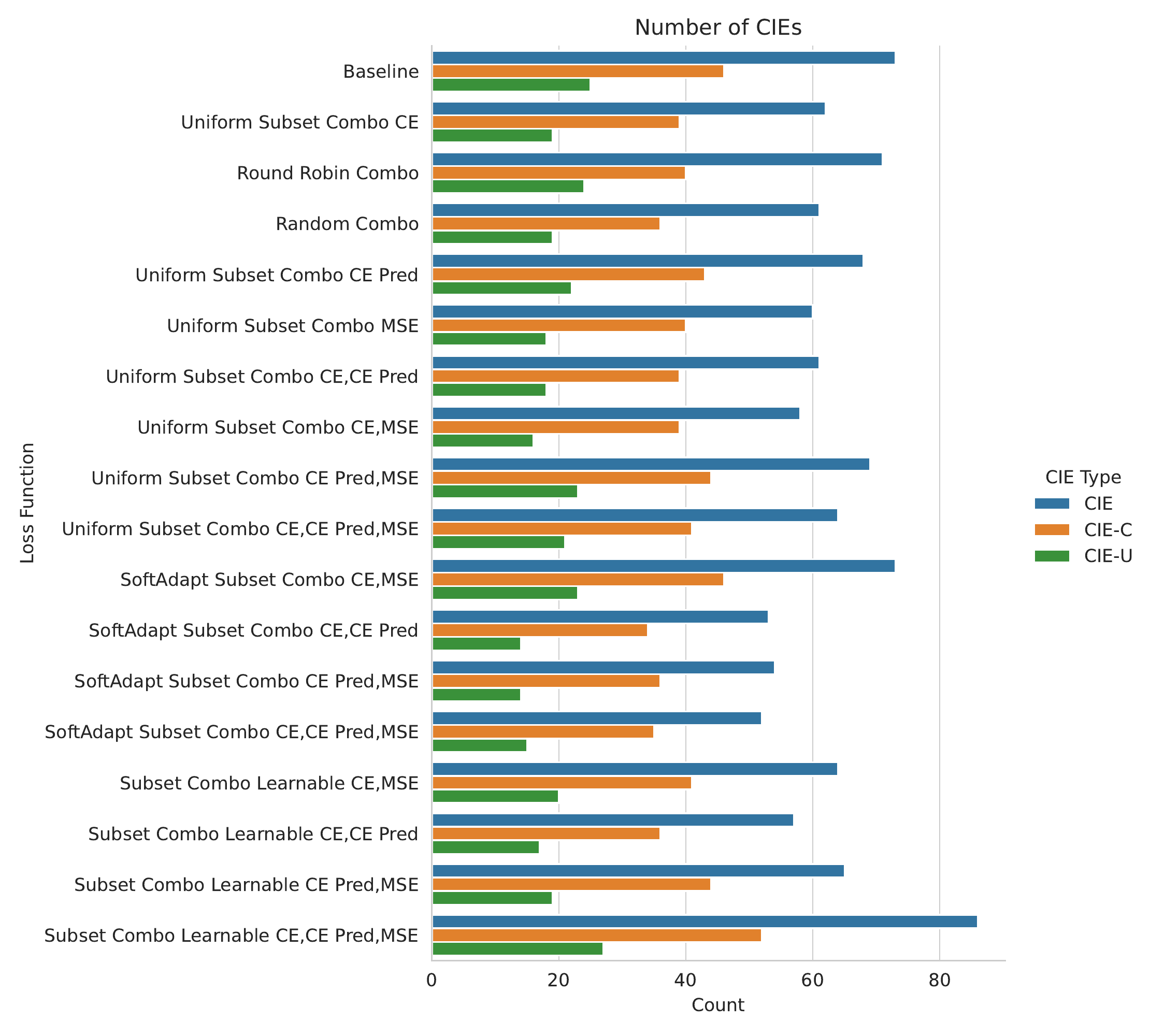}
    \caption{Number of CIEs resulting from compression.}
    \label{fig:ciescount}
\end{figure}



%% file: iclr2021_conference.bbl
\begin{thebibliography}{18}
\providecommand{\natexlab}[1]{#1}
\providecommand{\url}[1]{\texttt{#1}}
\expandafter\ifx\csname urlstyle\endcsname\relax
  \providecommand{\doi}[1]{doi: #1}\else
  \providecommand{\doi}{doi: \begingroup \urlstyle{rm}\Url}\fi

\bibitem[Bojarski et~al.(2016)Bojarski, Del~Testa, Dworakowski, Firner, Flepp,
  Goyal, Jackel, Monfort, Muller, Zhang, et~al.]{bojarski2016end}
Mariusz Bojarski, Davide Del~Testa, Daniel Dworakowski, Bernhard Firner, Beat
  Flepp, Prasoon Goyal, Lawrence~D Jackel, Mathew Monfort, Urs Muller, Jiakai
  Zhang, et~al.
\newblock End to end learning for self-driving cars.
\newblock \emph{arXiv preprint arXiv:1604.07316}, 2016.

\bibitem[Brown et~al.(2020)Brown, Mann, Ryder, Subbiah, Kaplan, Dhariwal,
  Neelakantan, Shyam, Sastry, Askell, Agarwal, Herbert-Voss, Krueger, Henighan,
  Child, Ramesh, Ziegler, Morris, Kusner, Loftus, Graesser, Narayanan, and
  Yosinski]{Brown2020LanguageAI}
Tom~B. Brown, Benjamin Mann, Nick Ryder, Melanie Subbiah, Jared Kaplan,
  Prafulla Dhariwal, Arvind Neelakantan, Pranav Shyam, Girish Sastry, Amanda
  Askell, Sandhini Agarwal, Ariel Herbert-Voss, Gretchen Krueger, Tom Henighan,
  Rewon Child, Aditya Ramesh, Daniel~M. Ziegler, Chris Morris, Matt Kusner,
  Joshua Loftus, Alex Graesser, Aswin~C. Narayanan, and Jason Yosinski.
\newblock Language models are few-shot learners.
\newblock \emph{arXiv preprint arXiv:2005.14165}, 2020.

\bibitem[Deng et~al.(2009)Deng, Dong, Socher, Li, Li, and Fei-Fei]{imagenet}
Jia Deng, Wei Dong, Richard Socher, Li-Jia Li, Kai Li, and Li~Fei-Fei.
\newblock Imagenet: A large-scale hierarchical image database.
\newblock In \emph{2009 IEEE Conference on Computer Vision and Pattern
  Recognition}, pp.\  248--255, 2009.
\newblock \doi{10.1109/CVPR.2009.5206848}.

\bibitem[Feldman(2020)]{feldman2020does}
Vitaly Feldman.
\newblock Does learning require memorization? a short tale about a long tail.
\newblock In \emph{Proceedings of the 52nd Annual ACM SIGACT Symposium on
  Theory of Computing}, pp.\  954--959, 2020.

\bibitem[Feldman \& Zhang(2020)Feldman and Zhang]{feldman2020neural}
Vitaly Feldman and Chiyuan Zhang.
\newblock What neural networks memorize and why: Discovering the long tail via
  influence estimation.
\newblock \emph{Advances in Neural Information Processing Systems},
  33:\penalty0 2881--2891, 2020.

\bibitem[Gouert et~al.(2023{\natexlab{a}})Gouert, Joseph, Dalton, Augonnet,
  Garland, and Tsoutsos]{gouert2023accelerated}
Charles Gouert, Vinu Joseph, Steven Dalton, Cedric Augonnet, Michael Garland,
  and Nektarios~Georgios Tsoutsos.
\newblock Accelerated encrypted execution of general-purpose applications.
\newblock \emph{Cryptology ePrint Archive}, 2023{\natexlab{a}}.

\bibitem[Gouert et~al.(2023{\natexlab{b}})Gouert, Joseph, Dalton, Augonnet,
  Garland, and Tsoutsos]{gouert2023arctyrex}
Charles Gouert, Vinu Joseph, Steven Dalton, Cedric Augonnet, Michael Garland,
  and Nektarios~Georgios Tsoutsos.
\newblock Arctyrex: Accelerated encrypted execution of general-purpose
  applications.
\newblock \emph{arXiv preprint arXiv:2306.11006}, 2023{\natexlab{b}}.

\bibitem[Heydari et~al.(2019)Heydari, Thompson, and Mehmood]{softadapt}
A.~Ali Heydari, Craig~A. Thompson, and Asif Mehmood.
\newblock Softadapt: Techniques for adaptive loss weighting of neural networks
  with multi-part loss functions, 2019.
\newblock URL \url{https://arxiv.org/abs/1912.12355}.

\bibitem[Hinton et~al.(2015)Hinton, Vinyals, and Dean]{hinton2015distilling}
Geoffrey Hinton, Oriol Vinyals, and Jeff Dean.
\newblock Distilling the knowledge in a neural network.
\newblock \emph{arXiv preprint arXiv:1503.02531}, 2015.

\bibitem[Hooker et~al.(2019)Hooker, Courville, Clark, Dauphin, and
  Frome]{hooker2019compressed}
Sara Hooker, Aaron Courville, Gregory Clark, Yann Dauphin, and Andrea Frome.
\newblock What do compressed deep neural networks forget?
\newblock \emph{arXiv preprint arXiv:1911.05248}, 2019.

\bibitem[Joseph(2021)]{DBLP:phd/us/Joseph21}
Vinu Joseph.
\newblock \emph{Programmable Neural Network Compression with Correctness
  Emphasis}.
\newblock PhD thesis, University of Utah, {USA}, 2021.

\bibitem[Joseph et~al.(2020{\natexlab{a}})Joseph, Chalapathi, Bhaskara,
  Gopalakrishnan, Panchekha, and Zhang]{joseph2020correctness}
Vinu Joseph, Nithin Chalapathi, Aditya Bhaskara, Ganesh Gopalakrishnan, Pavel
  Panchekha, and Mu~Zhang.
\newblock Correctness-preserving compression of datasets and neural network
  models.
\newblock In \emph{2020 IEEE/ACM 4th International Workshop on Software
  Correctness for HPC Applications (Correctness)}, pp.\  1--9. IEEE,
  2020{\natexlab{a}}.

\bibitem[Joseph et~al.(2020{\natexlab{b}})Joseph, Siddiqui, Bhaskara,
  Gopalakrishnan, Muralidharan, Garland, Ahmed, and Dengel]{joseph2020going}
Vinu Joseph, Shoaib~Ahmed Siddiqui, Aditya Bhaskara, Ganesh Gopalakrishnan,
  Saurav Muralidharan, Michael Garland, Sheraz Ahmed, and Andreas Dengel.
\newblock Going beyond classification accuracy metrics in model compression.
\newblock \emph{arXiv preprint arXiv:2012.01604}, 2020{\natexlab{b}}.

\bibitem[Krizhevsky et~al.(2009)Krizhevsky, Nair, and
  Hinton]{Krizhevsky09learningmultiple}
Alex Krizhevsky, Vinod Nair, and Geoffrey Hinton.
\newblock The cifar-10 dataset.
\newblock \url{https://www.cs.toronto.edu/~kriz/cifar.html}, 2009.

\bibitem[Lin et~al.(2017)Lin, Goyal, Girshick, He, and Dollár]{focalloss}
Tsung-Yi Lin, Priya Goyal, Ross Girshick, Kaiming He, and Piotr Dollár.
\newblock Focal loss for dense object detection, 2017.
\newblock URL \url{https://arxiv.org/abs/1708.02002}.

\bibitem[{OpenAI}(2021)]{OpenAI}
{OpenAI}.
\newblock {OpenAI Language Model}.
\newblock \url{https://openai.com/}, 2021.

\bibitem[Rav{\`\i} et~al.(2016)Rav{\`\i}, Wong, Deligianni, Berthelot,
  Andreu-Perez, Lo, and Yang]{ravi2016deep}
Daniele Rav{\`\i}, Charence Wong, Fani Deligianni, Melissa Berthelot, Javier
  Andreu-Perez, Benny Lo, and Guang-Zhong Yang.
\newblock Deep learning for health informatics.
\newblock \emph{IEEE journal of biomedical and health informatics}, 21\penalty0
  (1):\penalty0 4--21, 2016.

\bibitem[Tran \& Tran(2020)Tran and Tran]{tran2020deep}
Minh-Ngoc Tran and Dat Tran.
\newblock Deep neural networks for financial forecasting: A comparative study.
\newblock \emph{Neurocomputing}, 369:\penalty0 100--109, 2020.

\end{thebibliography}
